\documentclass[runningheads]{llncs}
\usepackage[T1]{fontenc}
\usepackage{graphicx}
\usepackage{algorithm} 
\usepackage{algpseudocode}
\usepackage{amsmath,amsfonts}

\begin{document}
\title{Zweistein: A Dynamic Programming Evaluation Function for Einstein Würfelt Nicht!}
%
%
\author{Wei-Lin Hsueh\inst{1}\orcidID{0009-0005-5325-2003} \and
Tsan-sheng Hsu\inst{2}\orcidID{0000-0001-9721-7617}}
\authorrunning{W. L. Hsueh, T. S. Hsu}
%
\institute{Department of Computer Science and Information Engineering, National Taiwan University, Taipei 106319, Taiwan (R.O.C.)\\
\email{b08902034@ntu.edu.tw} \and
Institute of Information Science, Academia Sinica, Taipei 115201, Taiwan (R.O.C.)
\email{tshsu@iis.sinica.edu.tw}\\
\url{https://homepage.iis.sinica.edu.tw/pages/tshsu}}
\maketitle              
\begin{abstract}
This paper introduces Zweistein, a dynamic programming evaluation function for Einstein Würfelt Nicht! (EWN). Instead of relying on human knowledge to craft an evaluation function, Zweistein uses a data-centric approach that eliminates the need for parameter tuning. The idea is to use a vector recording the distance to the corner of all pieces. This distance vector captures the essence of EWN. It not only outperforms many traditional EWN evaluation functions but also won first place in the TCGA 2023 competition.

\end{abstract}

\section{Introduction}
\begin{figure}[htbp]
    \begin{minipage}{0.3\textwidth}
    \centerline{\includegraphics[width=0.7\textwidth]{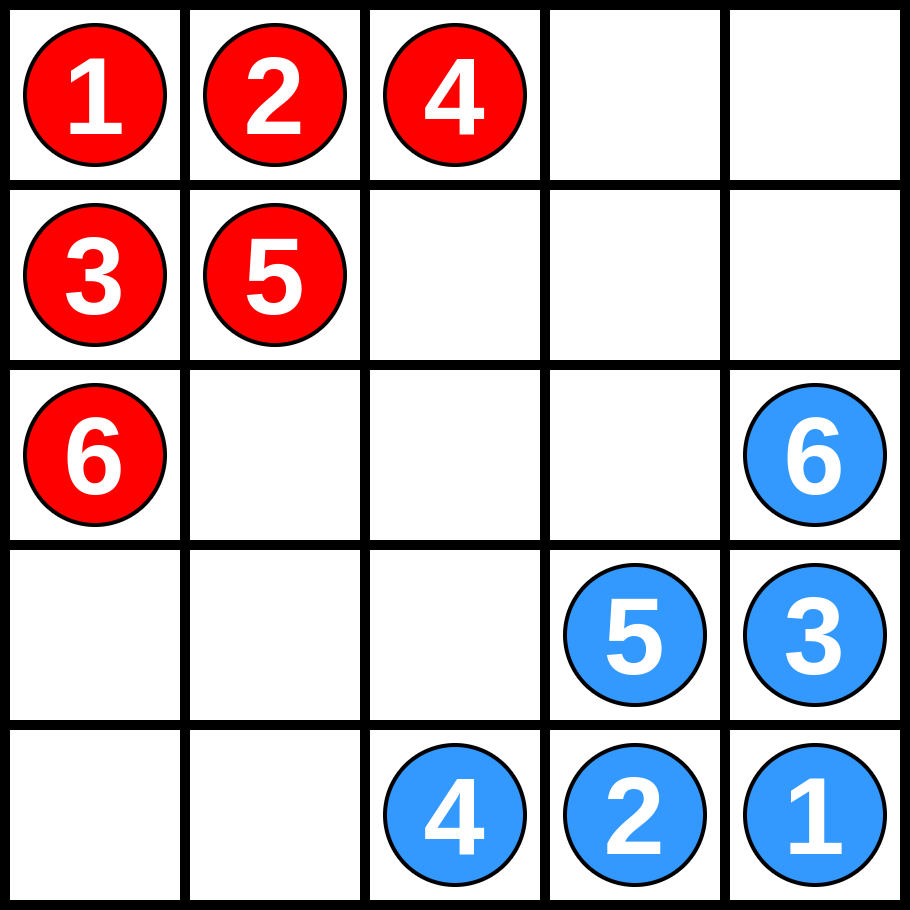}}
    \caption{The initial board.}
    \label{fig_init_board}
    \end{minipage}
    \begin{minipage}{0.3\textwidth}
    \centerline{\includegraphics[width=1\textwidth]{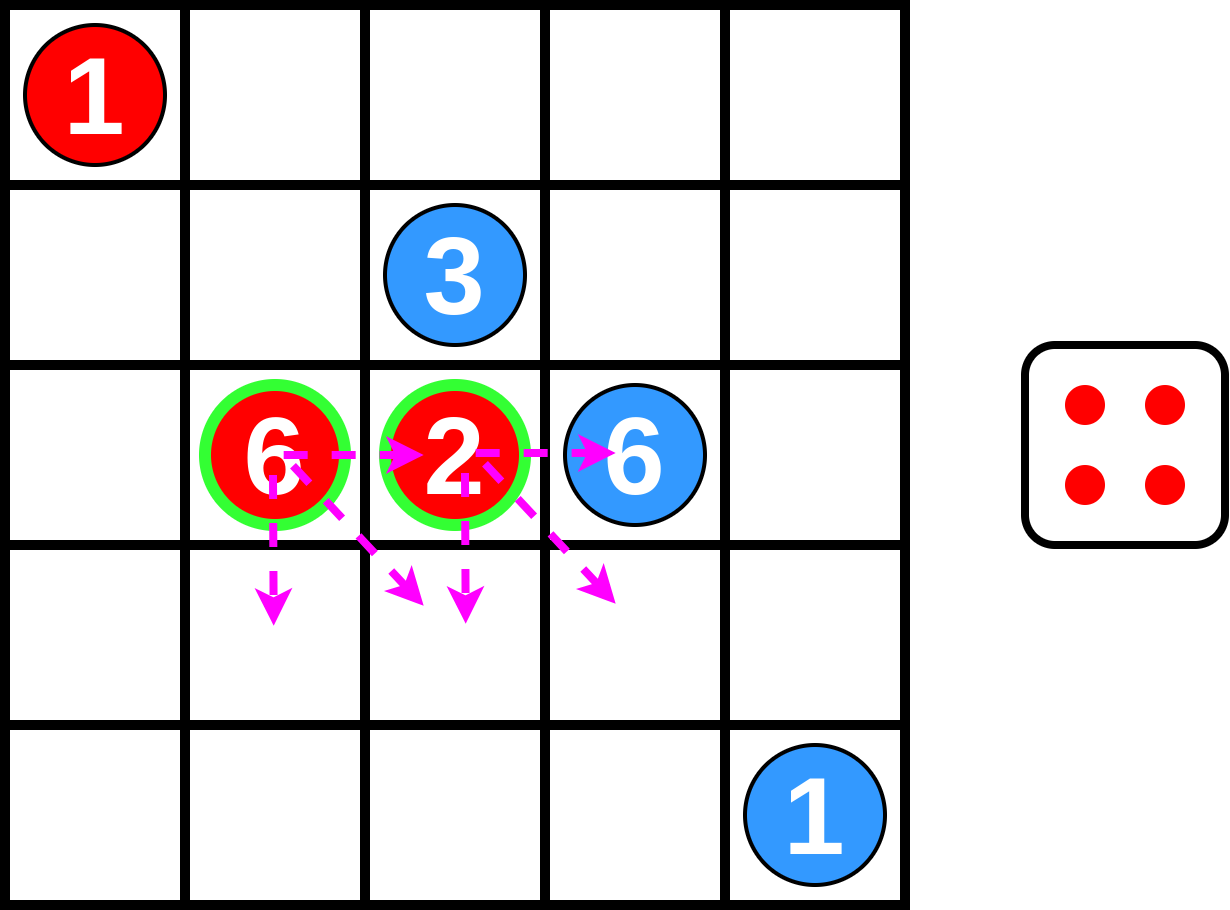}}
    \caption{An example showing the piece selection and movement}
    \label{fig_movement1}
    \end{minipage}
    \begin{minipage}{0.3\textwidth}
    \centerline{\includegraphics[width=1\textwidth]{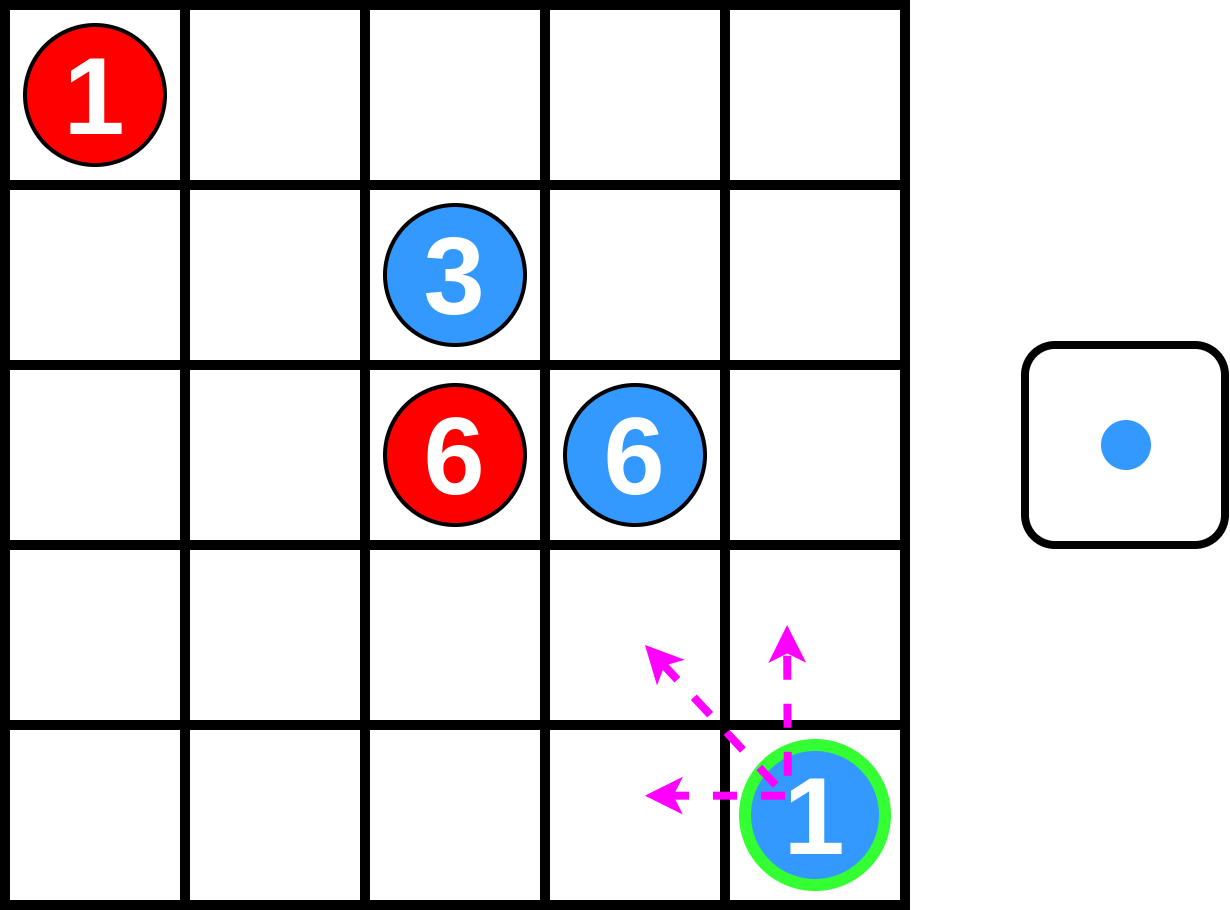}}
    \caption{Another example showing the piece selection and movement}
    \label{fig_movement2}
    \end{minipage}
\end{figure}

EinStein würfelt nicht! (abbr. EWN) is a two-player stochastic game with perfect information \cite{EWN_perfect_information}, played on a 5×5 board. Pieces are red or blue and labeled 1 to 6. Red starts in the top left, blue in the bottom right. The board's initial placement is usually symmetrical (see Fig.~\ref{fig_init_board}).

Players roll dice to move pieces. A piece can only move if its label matches the dice roll; otherwise, the player can move a piece with the next higher or lower number. Red pieces move right, down, or diagonally right-down, while blue pieces move left, up, or diagonally left-up. Pieces capture any piece at their destination, regardless of color.

Taking Fig.\ref{fig_movement1} as an example, the red side has pieces 1, 2, and 6. With a dice roll of 4, piece 4 is unavailable. Thus, the movable pieces are 6 (next higher) and 2 (next lower). Moving piece 6 to the right captures piece 2, as shown in Fig.\ref{fig_movement2}. In the next move, with a dice roll of 1, the blue side must move piece 1. The available moves for blue are also depicted in Fig.~\ref{fig_movement2}.

The game ends when one side captures all opponent pieces or a piece reaches the opponent's corner square (bottom right for red, top left for blue), with no draws possible.

\begin{figure}[htbp]
\centerline{\includegraphics[width=0.45\textwidth]{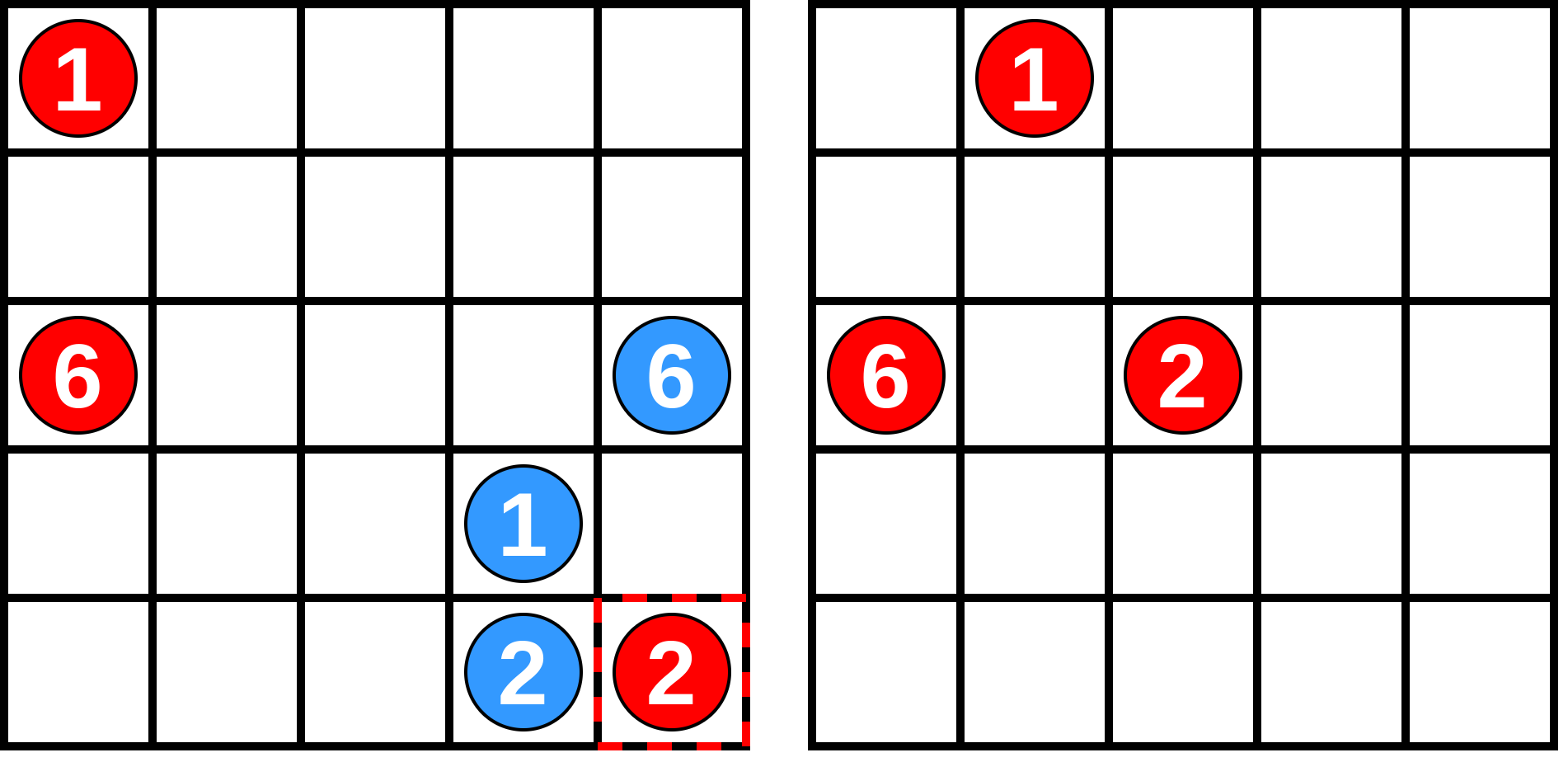}}
\caption{An example showing the win conditions. On the left side, red reaches the corner and wins. On the right side, red captures all blue pieces and wins.}
\label{fig}
\end{figure}

Despite the game's simple rules, designing an AI for EWN is challenging due to its inherent randomness and the strategic balance between defense and attack.

Traditional minimax tree search methods are less common in modern EWN contests. Instead, Monte Carlo tree search or machine learning techniques are more commonly used. However, this does not mean minimax search is inferior to other methods. This paper introduces a novel dynamic programming evaluation function that bypasses the need for human-crafted parameters, just like Monte Carlo methods or machine learning. Its effectiveness was demonstrated in the TCGA 2023 competition, where it secured first place against contestants who had previously won ICGA tournaments using machine learning and Monte Carlo methods \cite{monte_alpha}.

\section{Background}

\subsection{Motivation}
    Evaluation functions are essential in computer gaming programs. Claude E. Shannon recognized this as early as the 1940s \cite{shannon}, noting that an ideal evaluation function should correlate positively with the quality of a game position.

Achieving an ideal evaluation function, however, is equivalent to solving the game, which is infeasible for most popular games, including EWN. In practice, evaluation functions are approximations. Using Minimax search can reduce errors and strengthen the computer program.

Designing an effective evaluation function often requires extensive human knowledge, with no standardized method. For games like chess, abundant human insight makes creating a strong evaluation function easier, as demonstrated by IBM's Deep Blue \cite{deep blue}, which defeated top human players.

For newly invented or complex games with limited human knowledge, designing evaluation functions can be time-consuming. Go, for example, requires fine-tuning numerous parameters. While GnuGo \cite{gnugo} was a strong program in the early 2000s, it still could not defeat top players.

Recent methods, such as Monte Carlo tree search and reinforcement learning, minimize reliance on human knowledge. AlphaGo \cite{AlphaGo} used these techniques to beat the best Go players, showing that avoiding human expertise may be more effective for games with limited knowledge.

However, these methods do not provide insights that help players improve. Therefore, we introduce a dynamic programming approach to construct an evaluation function that requires no manual fine-tuning and offers a human-understandable heuristic.
    \subsection{Related works}
    EWN, being a game with only 20 years of history, has limited human knowledge available. Consequently, designing an evaluation function for EWN is challenging and complex. Most researchers have focused on using Monte Carlo tree search techniques to develop EWN programs \cite{EWN_perfect_information} \cite{TAAI_MCTS} \cite{china_MCTS} \cite{monte_alpha}.

    Several attempts have also been made to craft evaluation functions for EWN, often focusing on balancing attacking and defending power through parameter tuning \cite{EWNNY} \cite{ODEMA}.

    There was another evaluation function similar to Zweistein, called Schwarz table. \cite{Schwarz} In the experiments section, we will compare the performance of Zweistein with these existing functions.
    
\section{Methods}
This section primarily discusses the details of implementing Zweistein, which represents the win rate of EWN-simple, introduced below.
\subsection{EWN-simple}
    Zweistein employs a new concept from EWN-simple. While this new game resembles the original EWN rules, EWN-simple simplifies the win rate calculation significantly. The main difference between EWN-simple and vanilla EWN is the elimination of piece-capturing rules. Some adjustments were necessary to account for the removal of these rules.

    First, since EWN-simple does not allow piece capturing, two different pieces can overlap when they move to the same space.

\begin{figure}[htbp]
    \centerline{\includegraphics[width=0.5\textwidth]{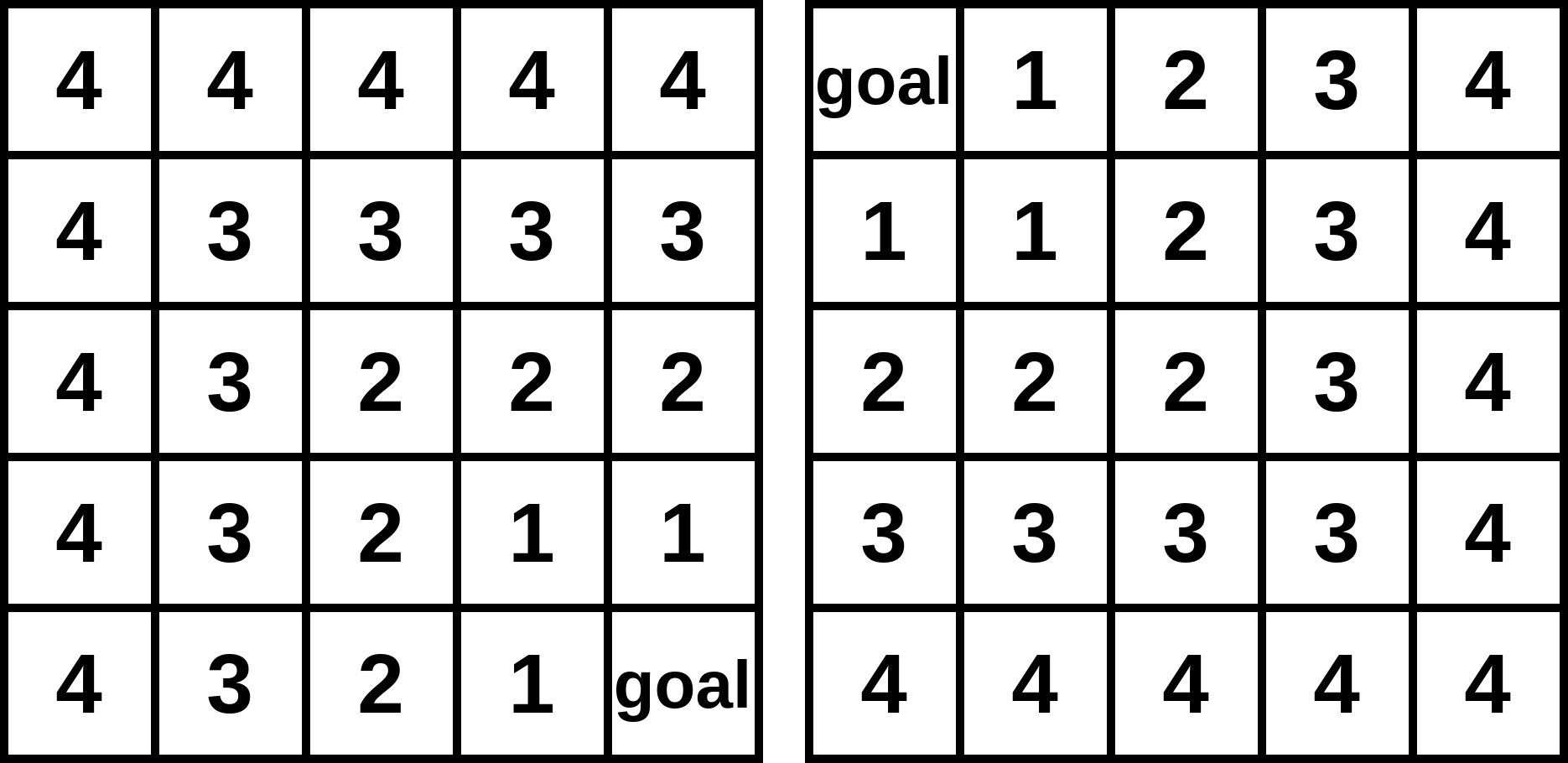}}
    \caption{The first table shows the Chebyshev distance to the goal corner from the red side's perspective. The second table provides a similar view from the blue side's perspective.}
    \label{fig_distance_to_corner}
    \end{figure}

    Second, EWN-simple introduces a special rule. Without the piece-capturing rule, each piece cannot interfere with others, there is no reason for a player to go along a longer path to the corner. Meaning players will always move their pieces toward the goal along the shortest path. To simplify the calculation, players are restricted to moving only along the shortest path, ensuring that the Chebyshev distance to the goal corner decreases by 1 with every move. The table of Chebyshev distance to the corner is shown in Fig.~\ref{fig_distance_to_corner}.

\subsection{Collapse an EWN-simple board into a distance array}
    This section is the core of Zweistein. The reason for removing the piece-capturing rules in EWN-simple is that it creates numerous isomorphic boards, allowing us to collapse all isomorphic boards into a single array form. This significantly reduces the space complexity.
    \begin{figure}[htbp]
        \begin{minipage}{0.5\textwidth}
        \centerline{\includegraphics[width=0.5\textwidth]{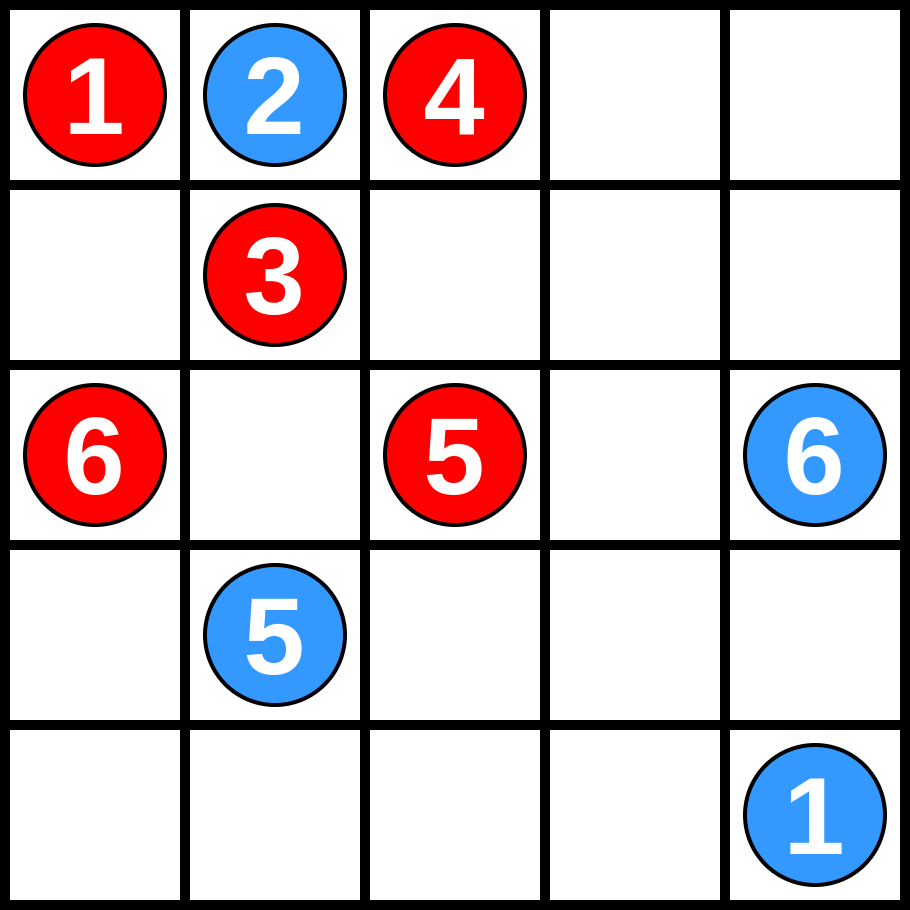}}
        \caption{An example board}
        \label{fig_example1}
        \end{minipage}
        \begin{minipage}{0.5\textwidth}
        \centerline{\includegraphics[width=0.5\textwidth]{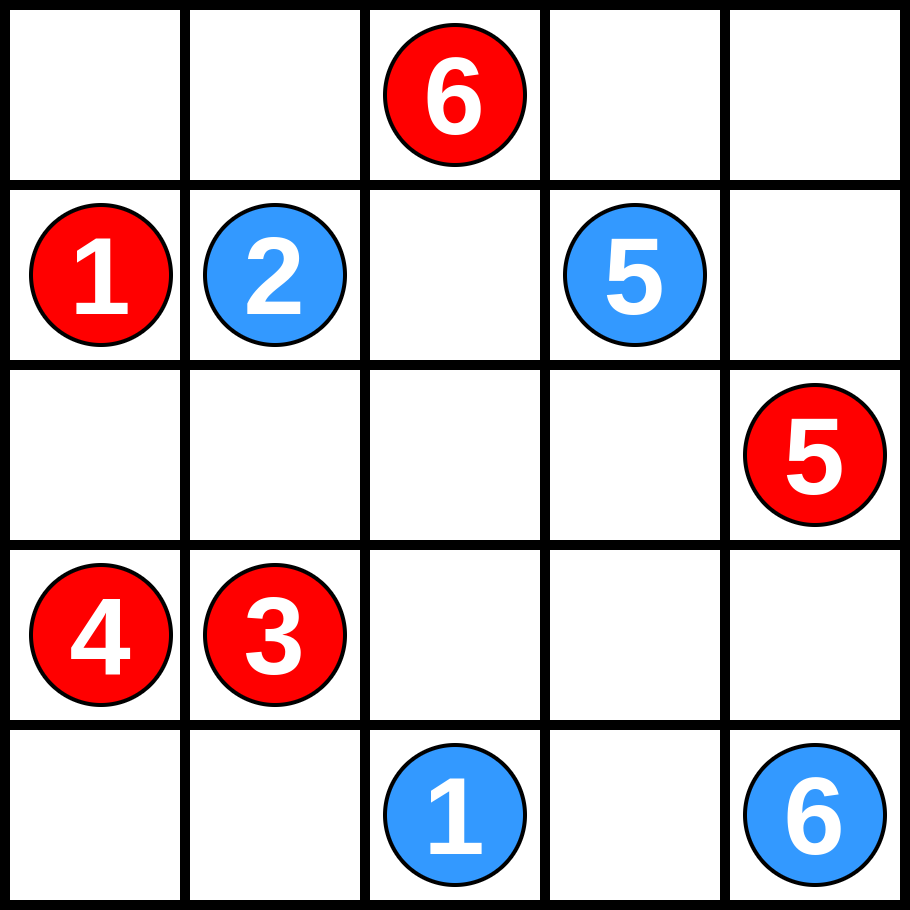}}
        \caption{An isomorphic board for Fig.~\ref{fig_example1}}
        \label{fig_example2}
        \end{minipage}
    \end{figure}
    
\begin{figure}[htbp]
    \centerline{\includegraphics[width=0.7\textwidth]{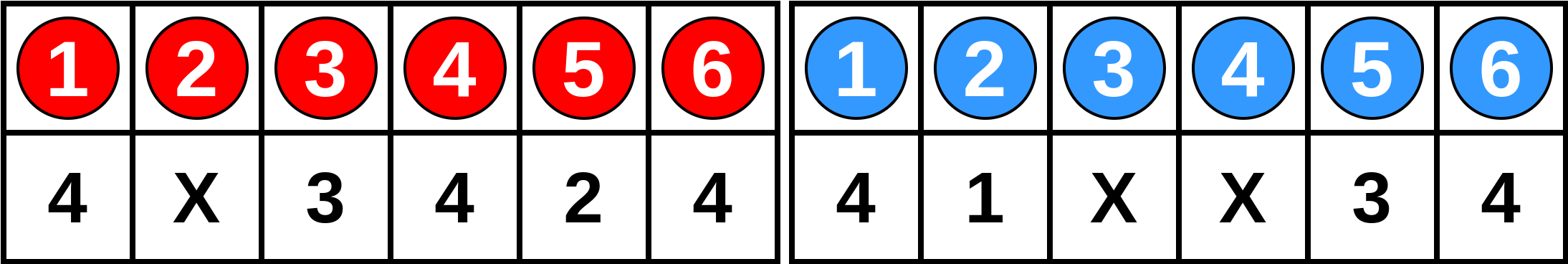}}
    \caption{The distance array for the Fig.~\ref{fig_example1} and Fig.~\ref{fig_example2}}
    \label{fig_array_form}
\end{figure}

Take Fig.~\ref{fig_example1} for example. We can use an array to describe the distance each piece has to the corner. For instance, according to Fig.~\ref{fig_distance_to_corner}, red piece 1 has a distance of 4, red piece 2 has been captured, red piece 3 has a distance of 3, and so on. Finally, the array for the red side and the blue side will look like Fig.~\ref{fig_array_form}. 

When collapsing the example in Fig.\ref{fig_example2} to an array, the result will be the same as the array shown in Fig.\ref{fig_array_form}. Therefore, Fig.\ref{fig_example1} and Fig.\ref{fig_example2} have the same distance array.

When two boards have the same distance array, they are considered isomorphic. In EWN-simple, each move decreases the distance to the corner by 1, meaning that pieces with the same distance to the corner exhibit the same behavior in playing.

\subsection{Building a database for EWN-simple}
    To save more space, Zweistein does not directly store the win rate for every possible game position. Instead, it employs some techniques to further reduce space complexity.
    \subsubsection{Viewing a player as a random variable}
    The first technique is to model a player as a random variable representing the number of steps required for any piece to reach the corner. To simplify this term, the number of steps required for any piece to reach the corner will be abbreviated as DTC (distance to corner).
    
    For example, let $X$ be a random variable that describes a player's behavior. The probability that the player's DTC is equal to $d$ is given by:
    \begin{equation}
     P(X=d)
    \end{equation}
    The range of DTC is [1, 19]. When DTC is 0, there is no need to calculate the evaluation function because the game has ended. The upper limit of 19 comes from the worst-case scenario: in this scenario, all 6 pieces initially have a distance of 4. They all move to a position where the distance is reduced to 1, which takes 18 steps. Finally, in the next step, one piece must reach the goal, adding up to a total of 19 steps.
    \subsubsection{Calculate Win rate of EWN-simple using DTC comparison}
    Because there is no piece-capturing rule in EWN-simple, the only way to win an EWN-simple game is to reach the corner. Therefore, the win rate of our side is the probability that our DTC is less than our opponent's DTC.
    
    Even if our side has the same DTC as the opponent, it is still considered a loss. When evaluating a game position, we have already completed our move, and the opponent will move next. Therefore, having the same DTC as the opponent is not sufficient for our side to win.

    To express this mathematically, let the DTC random variable of our side be denoted as $X$, and the DTC random variable of our opponent as $Y$. The probability that our DTC is less than our opponent's DTC can be written as:
    \begin{equation}
     P(X<Y)
    \end{equation}
    However, the value of this probability cannot be directly calculated. The formula should be split by the value of DTC, which ranges from 1 to 19. 
    \begin{equation}
    \begin{split}
     &P(X\leq0)P(Y=1)+P(X\leq1)P(Y=2)+\\
     &...+P(X\leq18)P(Y=19)\\
     \end{split}
    \end{equation}
    The summation form for (3) is thus:
    \begin{equation}
    \begin{split}
     &\sum_{i=1}^{19}P(X\leq i-1)P(Y=i)\\
    \end{split}
    \label{pdf_cdf_formula}
    \end{equation}

    Equation (\ref{pdf_cdf_formula}) shows that using a probability density function (abbr. pdf) and a cumulative density function (abbr. cdf) database of a DTC random variable is sufficient to calculate Zweistein, which represents the win rate in EWN-simple. The pseudo-code for utilizing the pdf and cdf database to compute Zweistein is provided in Algorithm \ref{winrate_algo}.
    \begin{algorithm}[htbp]
        \caption{Calculate Zweistein from pdf and cdf database}
        \begin{algorithmic}
        \Require pdf and cdf database: PDF\_VAL[15625][20],CDF\_VAL[15625][20]
        \Require our distance array and the opponent's distance array
        \State $our\_index$=ENCODE(our dist array)
        \State $oppo\_index$=ENCODE(opponent dist array)
        \State $sum=0$
        \For{$i=1,2,...,19$}
            \State $sum$+=CDF\_VAL[$our\_index$][i-1]*PDF\_VAL[$oppo\_index$][i]
        \EndFor
        \State \Return sum
        \end{algorithmic}
        \label{winrate_algo}
    \end{algorithm}

    As discussed in Section 3.2, calculating Zweistein requires collapsing the game into a distance array. Since there are only $5^6=15625$ different distance arrays for each side, we can use any encoding function to map a distance array to an index in the range $[0, 15624]$. This index can then be used to query the pdf and cdf database to determine the win rate.
    \subsubsection{Building the pdf database}
    Since constructing a cdf from a pdf is a straightforward programming task, this paper will focus on demonstrating how to build a DTC pdf database.

    Algorithm 2 is the pseudo-code for constructing the DTC pdf database. It follows a process similar to building an endgame database: perform a tree search for every game position and store the results after calculation. Because the database is very small, the entire building process takes less than a second to complete.

    There is a GitHub repository \cite{GitHub Zweistein} that implements this algorithm in C, available for anyone interested in using this evaluation function.

    \begin{algorithm}[htbp]
        \caption{pdf database generator}
        \begin{algorithmic}
        \Require a table that stores all results: PDF\_VAL[15625][20]
        \Procedure{CalculateExpectedValueOfDTC}{pdf}
        \State $sum$=0
        \For{$i=0,...,19$}
            \State $sum$ += $i$*pdf[$i$]
        \EndFor
        \State \Return $sum$
        \EndProcedure
        \Procedure{RightShift}{pdf}
        \State new\_pdf[20] = [0,...,0] (20 zeroes)
        \For{$i=0,1,2,...,19$}
        \State new\_pdf[$i$+1] = pdf[$i$]
        \EndFor
        \State \Return new\_pdf
        \EndProcedure
        \Procedure{Move}{dist array, $piece\ num$}
        \State dist array[$piece\ num$]-=1
        \State \Return dist array
        \EndProcedure
        \Procedure{TREE\_SEARCH}{dist array}
            \If {WIN CONDITION}
                \State \Return [1,0,...,0] (1 followed by 19 zeroes)
            \EndIf
            \State index=ENCODE(dist array)
            \If{PDF\_VAL[index] is visited}
                \State \Return PDF\_VAL[index]
            \Else
                \State sum=[0,...,0] (20 zeroes)
                \For {$dice=1,2,\ldots,6$}
                    \State $min\_exp=\infty$,min\_pdf=[]
                    \For{$num = \text{all movable piece number}$}
                        \State child array=Move(a copy of dist array, $num$)
                        \State child\_pdf=TREE\_SEARCH(child array)
                        \State $expval$=CalculateExpectedValueOfDTC(child\_pdf)
                        \If{$expval<min\_exp$}
                            \State $min\_exp=expval$
                            \State min\_pdf=child pdf
                        \EndIf
                    \EndFor
                    \State $sum$+=min\_pdf
                \EndFor
                \State PDF\_VAL[index]=RightShift(sum/6)
                \State mark PDF\_VAL[index] as visited
            \EndIf
        \EndProcedure
        \For {$index=1,2,\ldots,15624$}
		\State dist array=DECODE(index)
            \State TREE\_SEARCH(dist array)
	\EndFor
        \State \Return PDF\_VAL
        \end{algorithmic}
    \end{algorithm}

\section{Experiments}
    \subsection{Comparison with other functions}
    In Section 2.2, we mentioned some previous attempts to craft an evaluation function for EWN. In this section, we will compare Zweistein with those earlier approaches.

    All of the experiments were conducted on a computer equipped with dual AMD EPYC 9354 32-core processors, with the program written in C++. For each searching depth, Zweistein played with the other function 100,000 times.
    
    The first function we will compare is called ODEMA \cite{ODEMA}. This function combines the attacking power, threat power, and blocking power with the formula:
    \begin{equation}
    attack\_factor \times Attack + Block - threat\_factor \times Threat
    \end{equation}
    For our experiment, we used the parameters provided in the original paper: \newline $attack\_factor = 2.5$, $threat\_factor = 0.05$. Although the value of parameter $N$ was not specified in the paper, it is necessary to calculate $Block$. So we use $N=2$ for the experiment.
    
    \begin{table}[htbp]
    \caption{The win rate of Zweistein when playing with ODEMA}
        \centering
        \begin{tabular}{|l|c|c|c|c|c|c|c|}
        \hline
          searching depth & \textbf{1} & \textbf{2} & \textbf{3} & \textbf{4} & \textbf{5} & \textbf{6} & \textbf{7}\\
         \hline        Win rate of Zweistein &0.47284&0.59492&0.55789&0.62252&0.53955&0.60491&0.52779\\
         \hline
        \end{tabular}
    \end{table}

    The second function is called EWNNY \cite{EWNNY}. This function is very complicated and has many parameters, so we used the program written by the author directly for the experiment.

    \begin{table}[htbp]
    \caption{The win rate of Zweistein when playing with EWNNY}
        \centering
        \begin{tabular}{|l|c|c|c|c|c|c|c|}
        \hline
          search depth & \textbf{1} & \textbf{2} & \textbf{3} & \textbf{4} & \textbf{5} & \textbf{6} & \textbf{7}\\
         \hline        Win rate of Zweistein &0.38912&0.48866&0.45690&0.50192&0.48137&0.49841&0.49362\\
         \hline
        \end{tabular}
    \end{table}

    In Table 1, Zweistein appears weak during a depth-1 search, primarily due to the absence of piece-capturing rules. However, since piece captures are possible during the tree search, this process naturally compensates for Zweistein's blind spot. As a result, the function's performance improves rapidly with increasing search depth when playing against ODEMA. The win rate never drops below $50$\% when the depth is greater than 1. Notably, ODEMA only showed better performance when the search depth was an odd number, indicating that this function lacks stable performance across different search depths.

    In the second table, Zweistein's performance closely mirrors that of EWNNY when the search depth is greater than 4. Despite the similarity in performance, Zweistein stands out for being simpler and faster, with a speed advantage of $52.4$\% over EWNNY. Table 3 shows that Zweistein takes an average of 16.53 ns per call, while EWNNY takes an average of 25.20 ns.

    \begin{table}[htbp]
    \caption{Time analysis between Zweistein and EWNNY on AMD EPYC 9354}
        \centering
        \begin{tabular}{|l|c|c|}
        \hline
          function type & \textbf{Zweistein} & \textbf{EWNNY}\\
         \hline        Time spend after executing $10^8$ times &1.653 s&2.520 s\\
         \hline
        \end{tabular}
    \end{table}
    
    Although Zweistein is faster than EWNNY, it is still insufficient for an additional search layer. However, the increased speed provides greater flexibility in time management, making Zweistein stronger than EWNNY when the total time is fixed.

    An experiment was conducted to support this statement. We used dynamic time control to set the time limit for each move. The dynamic time control formula we used is as follows:
    
    \begin{equation}
    time\_limit\_each\_step = \frac{remaining\_time}{max((15-number\_of\_steps\_taken),3)}
    \end{equation}
    
    Table 4 shows the win rate between EWNNY and Zweistein at various fixed total times. Each cell represents the win rate over 100,000 games. We observe that when the total time exceeds 5 seconds, Zweistein achieves a higher win rate than EWNNY.

    \begin{table}[htbp]
    \caption{The win rate of Zweistein when playing with EWNNY in various fixed total times}
        \centering
        \begin{tabular}{|l|c|c|c|c|c|}
        \hline
          total time (in seconds) & \textbf{1} & \textbf{2} & \textbf{3} & \textbf{4} & \textbf{5} \\
         \hline        Win rate of Zweistein &0.496050&0.499440&0.499060&0.498910&0.502150\\
         \hline
          total time (in seconds) & \textbf{6} & \textbf{7} & \textbf{8} & \textbf{9} & \textbf{10}\\
         \hline        Win rate of Zweistein & 0.502710&0.501050&0.500350&0.501000&0.502430\\
         \hline
        \end{tabular}
    \end{table}
    \subsection{Comparison with exact win rate}
\begin{figure}[htbp]
        \begin{minipage}{0.5\textwidth}
            \centering
            \centerline{\includegraphics[width=0.5\textwidth]{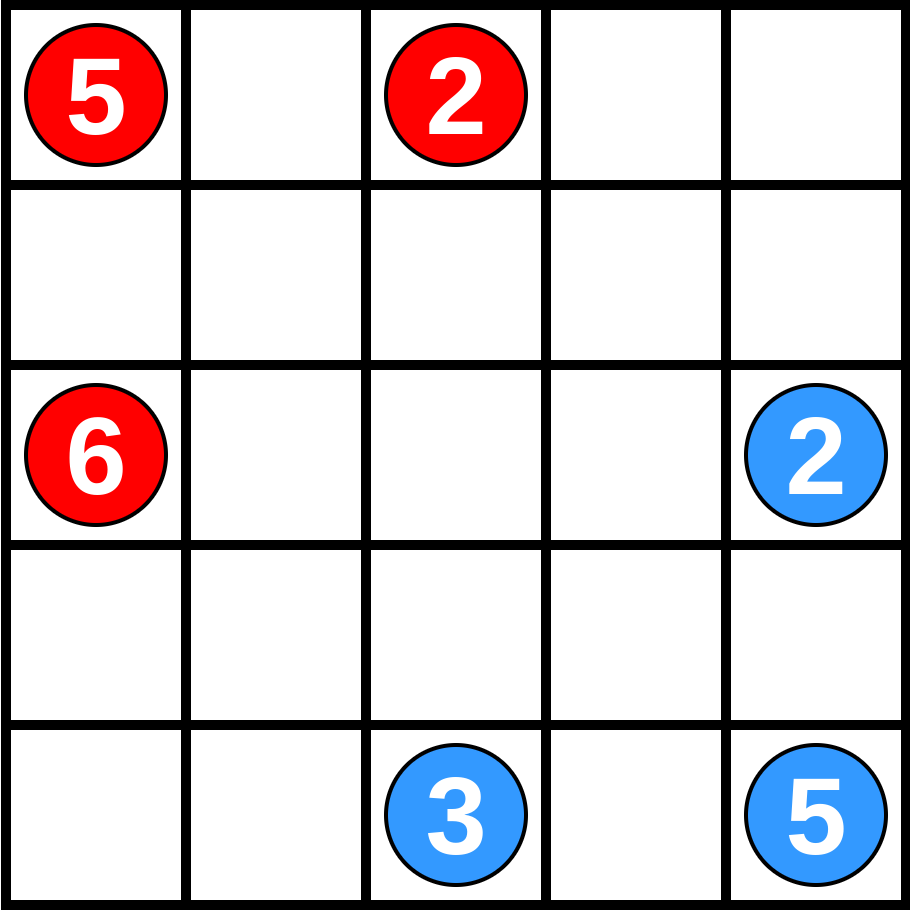}} 
            \caption{An example board that Zweistein \newline is accurate}
            \label{fig_win_rate_example1}
        \end{minipage}
        \begin{minipage}{0.5\textwidth}
            \centering
            \centerline{\includegraphics[width=0.5\textwidth]{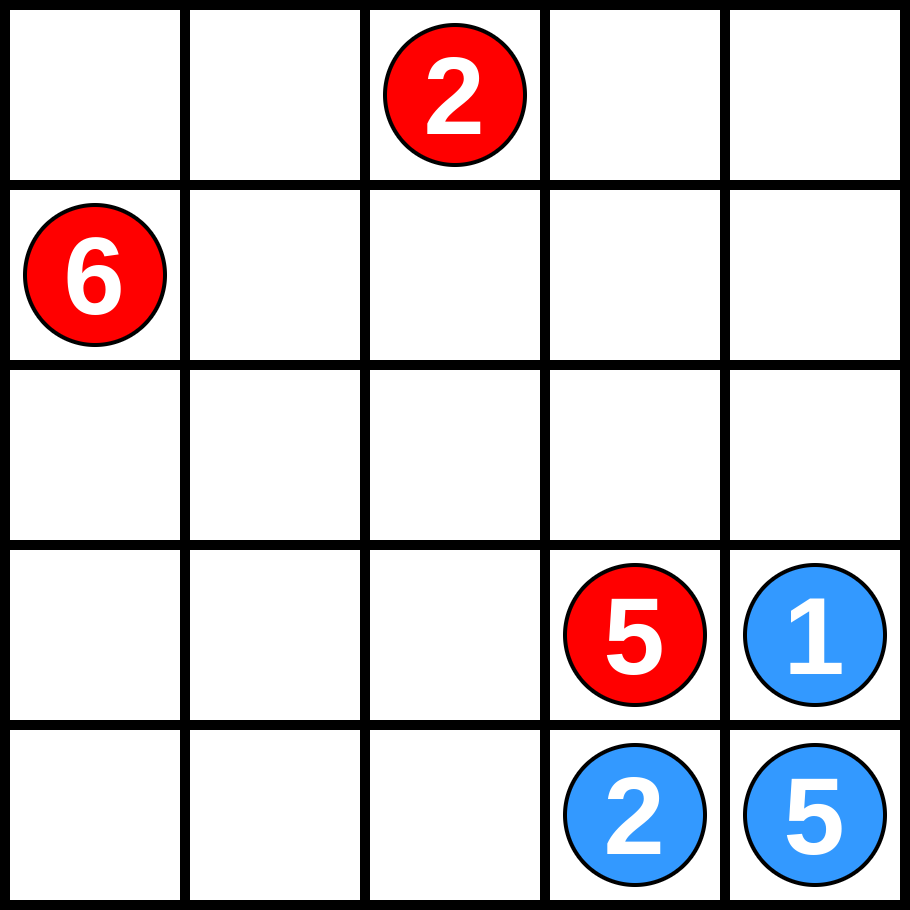}} 
            \caption{Another example board that Zweistein is not accurate}
            \label{fig_win_rate_example2}
        \end{minipage}
    \end{figure}

    In this section, we use simple boards for which the exact win rate can be calculated by brute force. We then compare the exact win rate and evaluation values of Zweistein.
    Since Zweistein does not include piece-capturing rules, when each piece is separated from the others and capture is not possible on the next move, Zweistein's evaluation value closely approximates the exact win rate. For example, in Fig.\ref{fig_win_rate_example1}, if the blue side moves first, the exact win rate for the red side is $0.564142$, while Zweistein's value is $0.588791$.

    On the other hand, when piece-capturing has a significant impact on the win rate, Zweistein's value becomes inaccurate. For example, in Fig.\ref{fig_win_rate_example2}, the blue side can avoid losing the game by capturing red 2, but Zweistein does not take this into consideration. If the blue side moves first, the exact win rate for the red side is $0.659595$, while Zweistein's value is $0.954475$. 
    
    However, by performing a depth-1 search, the influence of piece capture can be detected by the search algorithm, making the depth-1 Zweistein value $0.608869$, which is more accurate than the original version. This is the primary reason why Zweistein performs poorly at lower search depths but performs better at higher search depths.
    
    \subsection{Comparison with Schwarz table}
    A 2005 student study report introduced a similar evaluation function called the Schwarz Table.\cite{Schwarz} Instead of calculating the win rate of EWN-simple, the expected value of DTC was calculated.
    In this subsection, we will compare the value of Zweistein with that of the Schwarz Table.

\begin{figure}[htbp]
        \begin{minipage}{0.5\textwidth}
            \centering
            \centerline{\includegraphics[width=0.5\textwidth]{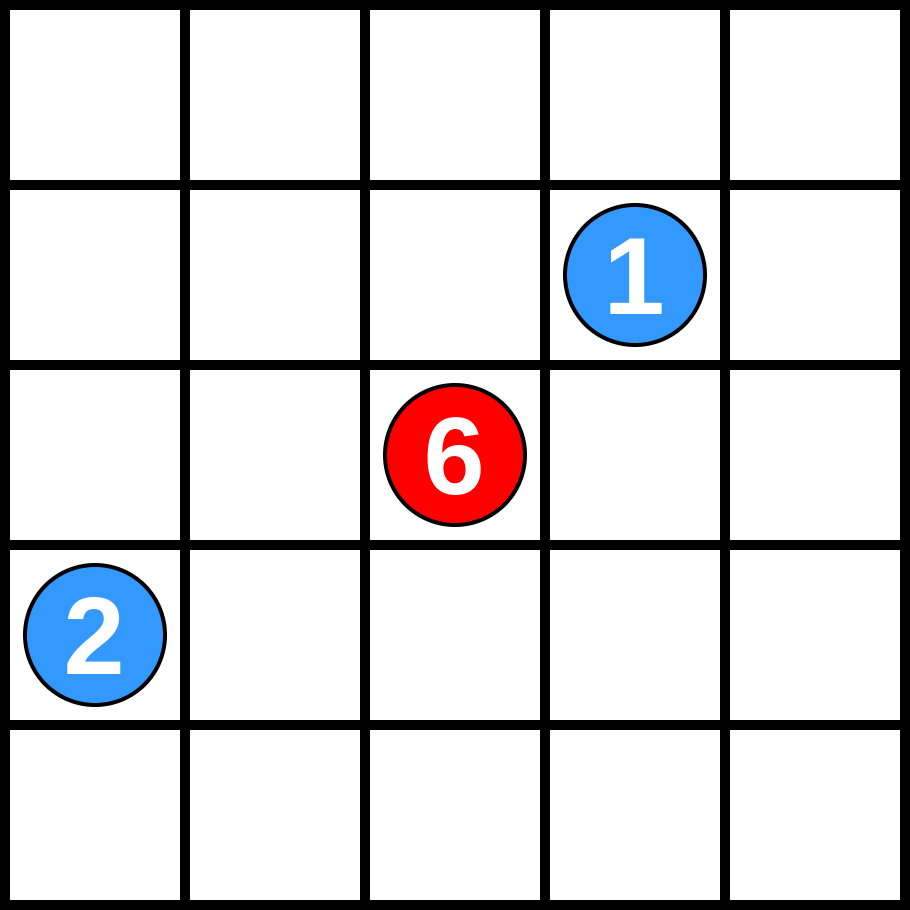}} 
            \caption{First example, guaranteed win \newline for red.}
            \label{schwarz_example1}
        \end{minipage}
        \begin{minipage}{0.5\textwidth}
            \centering
            \centerline{\includegraphics[width=0.5\textwidth]{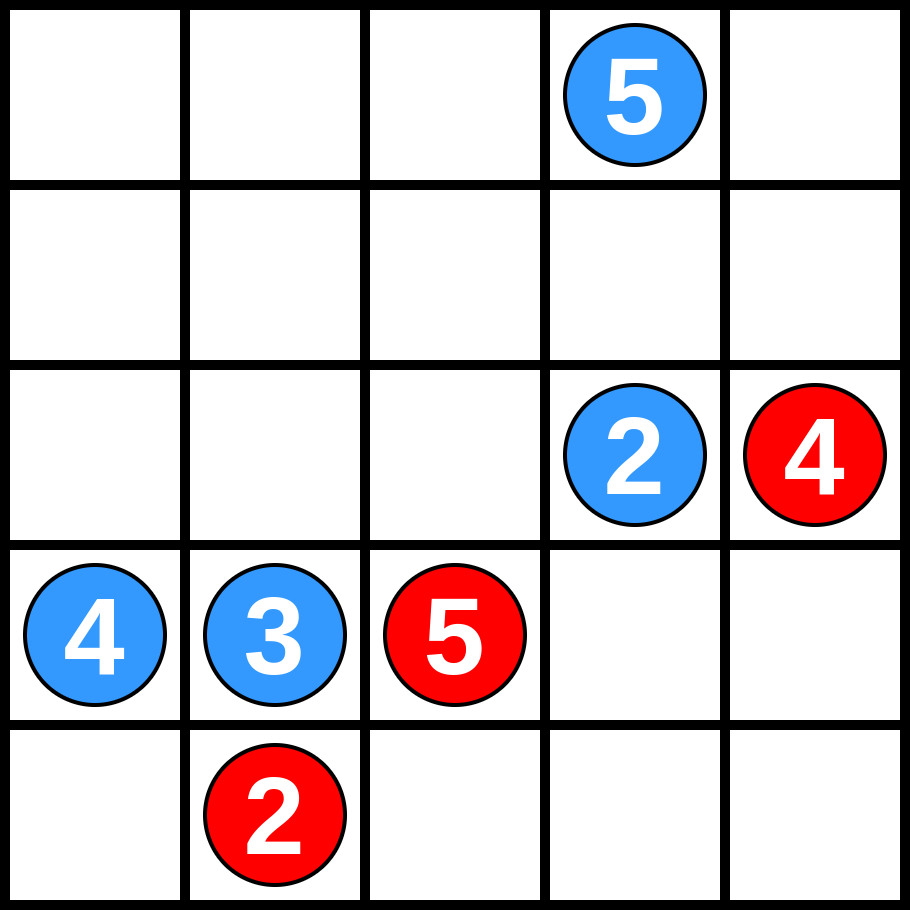}} 
            \caption{Second example, not a guaranteed win for red.}
            \label{schwarz_example2}
        \end{minipage}
    \end{figure}

    In Fig.\ref{schwarz_example1}, it is a guaranteed win for red when blue moves first. The value of Zweistein is 1, while the value of Schwarz table is $3.532407 - 2.000000 = 1.532407$.

    In Fig.\ref{schwarz_example2}, the win rate for red is 0.829276.
    The value of Zweistein is 0.837577, while the value of Schwarz table is $5.581104 - 3.305556 = 2.275548$.

    Therefore, Zweistein is an improved version of Schwarz Table, as the latter sometimes underestimates strong board positions (e.g., Fig.\ref{schwarz_example1}) or overestimates less favorable ones (e.g., Fig.\ref{schwarz_example2}).
    
    \section{Computer tournaments}
    Zweistein participated in the TCGA 2023 tournament and secured first place among 8 contestants. Table 5 provides the contest record, demonstrating the strength and effectiveness of this evaluation function.
    \begin{table}[htbp]
    \caption{The record of TCGA2023}
        \centering
        \scriptsize
        \begin{tabular}{|l|l|c|c|c|c|c|c|c|c|c|c|c|c|c|c|c|c|}
        \hline
        \textbf{No.} & \textbf{rounds} &  \multicolumn{2}{c|}{\textbf{1}} & \multicolumn{2}{c|}{\textbf{2}} & \multicolumn{2}{c|}{\textbf{3}} & \multicolumn{2}{c|}{\textbf{4}} & \multicolumn{2}{c|}{\textbf{5}} & \multicolumn{2}{c|}{\textbf{6}} & \multicolumn{2}{c|}{\textbf{7}} & \multicolumn{2}{c|}{\textbf{result}} \\
        \hline
        & & \textbf{\textit{opp}} & \textbf{\textit{win}} & \textbf{\textit{opp}} & \textbf{\textit{win}} & \textbf{\textit{opp}} & \textbf{\textit{win}} & \textbf{\textit{opp}} & \textbf{\textit{win}} & \textbf{\textit{opp}} & \textbf{\textit{win}} & \textbf{\textit{opp}} & \textbf{\textit{win}} & \textbf{\textit{opp}} & \textbf{\textit{win}} & \textbf{\textit{sum}} & \textbf{\textit{rank}} \\
        \hline
        1 & EWN\_AI & 6 & 37 & 2 & 39 & 3 & 59 & 4 & 50 & 5 & 68 & 8 & 76 & 7 & 53 & 382 & 4 \\
        \hline
        2 & Zweistein & 5 & 57 & 1 & 61 & 7 & 53 & 6 & 53 & 8 & 83 & 3 & 73 & 4 & 49 & 429 & 1 \\
        \hline
        3 & Reinstein & 4 & 35 & 5 & 48 & 1 & 41 & 7 & 33 & 6 & 32 & 2 & 27 & 8 & 74 & 290 & 7 \\
        \hline
        4 & Monte\_Alpha & 3 & 65 & 8 & 73 & 5 & 66 & 1 & 50 & 7 & 50 & 6 & 46 & 2 & 51 & 401 & 3 \\
        \hline
        5 & EWN\_Alpha & 2 & 43 & 3 & 52 & 4 & 34 & 8 & 62 & 1 & 32 & 7 & 46 & 6 & 38 & 307 & 6 \\
        \hline
        6 & deku\_Ein & 1 & 63 & 7 & 52 & 8 & 68 & 2 & 47 & 3 & 68 & 4 & 54 & 5 & 62 & 414 & 2 \\
        \hline
        7 & ssunoo & 8 & 66 & 6 & 48 & 2 & 47 & 3 & 67 & 4 & 50 & 5 & 54 & 1 & 47 & 379 & 5 \\
        \hline
        8 & MuMu & 7 & 34 & 4 & 27 & 6 & 32 & 5 & 38 & 2 & 17 & 1 & 24 & 3 & 26 & 198 & 8 \\
        \hline
        \end{tabular}
\end{table}

    \section{Concluding remarks and future work}
    In this paper, we introduced a novel approach to designing an EWN evaluation function by simplifying the game rules and constructing a database based on these simplified rules. This resulted in a simple, yet powerful, evaluation function that requires no parameter tuning. It also acts as a strong and simple baseline when crafting a stronger EWN evaluation function.

    Although experiments demonstrate that it is possible to construct an evaluation function with performance similar to Zweistein using traditional methods, these functions typically require parameter tuning and are significantly more computationally expensive to develop. Additionally, the potential for improving traditional evaluation functions is limited. Due to the curse of dimensionality \cite{curse of dim}, adding a new parameter to the function exponentially multiplies the time complexity of grid search.
    
    For example, EWNNY \cite{EWNNY} requires an extensive grid search to acquire an optimal set of parameters, with each set of parameters necessitating a $100,000$-round experiment.
    
    The current version of Zweistein requires only a few megabytes of space, which is relatively small by today’s standards. This suggests substantial potential for further development and optimization. Expanding the database to include partial rules for piece capturing presents a promising direction for enhancing Zweistein's performance.

\begin{credits}
\subsubsection{\ackname}
This research was partially supported by the Ministry of Science and Technology Council (NSTC) of Taiwan under grant numbers 111-2221-E-001-017-MY3. We would also like to express our gratitude to the reviewers for their valuable feedback.
\subsubsection{\discintname}
The authors have no competing interests. 
\end{credits}

\end{document}